\documentclass[sigconf]{acmart}

\usepackage{booktabs} % For formal tables

\copyrightyear{2019}
\acmYear{2019}
\setcopyright{rightsretained}
\acmConference[SAC '19]{The 34th ACM/SIGAPP Symposium on Applied Computing}{April 8--12, 2019}{Limassol, Cyprus}
\acmBooktitle{The 34th ACM/SIGAPP Symposium on Applied Computing (SAC '19), April 8--12, 2019, Limassol, Cyprus}
\acmDOI{10.1145/3297280.3297586}
\acmISBN{978-1-4503-5933-7/19/04}

\usepackage{stfloats}
\usepackage{hyperref}
\usepackage{eurosym}
\usepackage{graphicx}
\usepackage{float}
\usepackage{booktabs}
\usepackage{footnote}	
\usepackage{wrapfig}
\usepackage{subcaption}
\usepackage{caption}
\usepackage[section]{placeins}
\usepackage{multirow}

\begin{document}
\title{Multiple perspectives HMM-based feature engineering for credit
card fraud detection}

\author{Yvan Lucas}
\affiliation{%
  \institution{INSA Lyon \and Universit{\"a}t Passau}}
\email{yvanlucas44@gmail.com}

\author{Pierre-Edouard Portier}
\affiliation{%
  \institution{INSA Lyon}}
\email{pierre-edouard.portier@insa-lyon.fr}

\author{L\'ea Laporte}
\affiliation{%
  \institution{INSA Lyon}}
\email{lea.laporte@insa-lyon.fr}

\author{Sylvie Calabretto}
\affiliation{%
  \institution{INSA Lyon}}
\email{sylvie.calabretto@insa-lyon.fr}

\author{Olivier Caelen}
\affiliation{%
  \institution{ATOS Worldline}}
\email{o.caelen@gmail.com}

\author{Liyun He-Guelton}
\affiliation{%
  \institution{ATOS Worldline}}
\email{liyun.he-guelton@worldline.com}

\author{Michael Granitzer}
\affiliation{%
  \institution{Universit{\"a}t Passau}}
\email{michael.granitzer@uni-passau.de}

% The default list of authors is too long for headers}
%
\renewcommand{\shortauthors}{Y. Lucas et al.}

\begin{abstract}
Machine learning and data mining techniques have been used extensively in order to detect credit card frauds. However, most studies consider credit card transactions as isolated events and not as a sequence of transactions.

In this article, we model a sequence of credit card transactions from three different perspectives, namely (i) does the sequence contain a Fraud? (ii) Is the sequence obtained by fixing the card-holder or the payment terminal? (iii) Is it a sequence of spent amount or of elapsed time between the current and previous transactions? Combinations of the three binary perspectives give eight sets of sequences from the (training) set of transactions. Each one of these sets is modelled with a Hidden Markov Model (HMM). Each HMM associates a likelihood to a transaction given its sequence of previous transactions. These likelihoods are used as additional features in a Random Forest classifier for fraud detection. This multiple perspectives HMM-based approach enables an automatic feature engineering in order to model the sequential properties of the dataset with respect to the classification task. This strategy allows for a 15\% increase in the precision-recall AUC compared to the state of the art feature engineering strategy for credit card fraud detection.% In addition, we study the repercussions of the addition of the proposed HMM-based features by looking at the importance of the features used by the Random Forest classifiers.

%The proposed feature engineering strategy opens perspectives for any supervised classification task with sequential datasets. To ensure reproducibility, an optimized code of the proposed framework can be found at \url{https://gitlab.com/Yvan_Lucas/hmm-ccfd}
\end{abstract}

%
% The code below should be generated by the tool at
% http://dl.acm.org/ccs.cfm
% Please copy and paste the code instead of the example below. 
%


\maketitle
\section{Introduction}

Credit card fraud detection presents several difficulties. One of them is the fact that the feature set describing a credit card transaction usually ignores detailed sequential information. Typical models only use raw transactional features, such as time, amount, merchant category, etc. Bolton \& al. \cite{bolton2001} showed the necessity to use attributes describing the history of the transaction when they used unsupervised methods such as peer group analysis for credit card fraud detection. Consequently, Whitrow \& al. \cite{whitrow2008} create descriptive statistics as features in order to include historical knowledge. These descriptive features can be for example the number of transactions or the total amount spent from the card-holder in the past 24 hours with the same merchant category or country. Among other authors, Bahnsen \& al. \cite{bahnsen2016} established Whitrow's transaction aggregation strategy as the state of the art feature engineering technique for credit card fraud detection. 

We identified several weaknesses in the construction of these features that motivated our work: Descriptive statistics provide an aggregated view over a set of transactions. Such aggregated features do not consider fine-grained temporal dependencies between the transactions. For example, a common fraud pattern starts with low amount transactions for testing the card, followed by high amount transaction. Moreover, these aggregated features consider only the history of the card-holder and do not exploit information of fraudulent transactions for feature engineering. However, a sequence of transactions happening at a fixed terminal can also contain valuable patterns for fraud detection.%Second, aggregated features are usually calculated over transactions following a fixed time window (e.g. 24 h). Transaction of very different card holders do not follow such a time pattern in general. Some cardholders have several transactions in the period, while others only have one transactions. Fixed size aggregated statistics can not account for that fact. Third, these features consider only the history of the card-holder and do not exploit information of fraudulent transactions for feature engineering. However, a sequence of transactions happening at a fixed terminal can also contain valuable patterns for fraud detection.

In our work we propose to generate history-based features using Hidden Markov models. These features are created by estimating the likelihood of a sequence of transactions to be regular in regard to terminal or cardholder transactions. More precisely, they quantify the similarity between an observed sequence and the sequence of past fraudulent or genuine transactions observed for the cardholders or the terminals.

\section{Multiple perspective HMM-based feature engineering}
\begin{figure*}[t]
\centerline{
\includegraphics[width=0.7\textwidth]{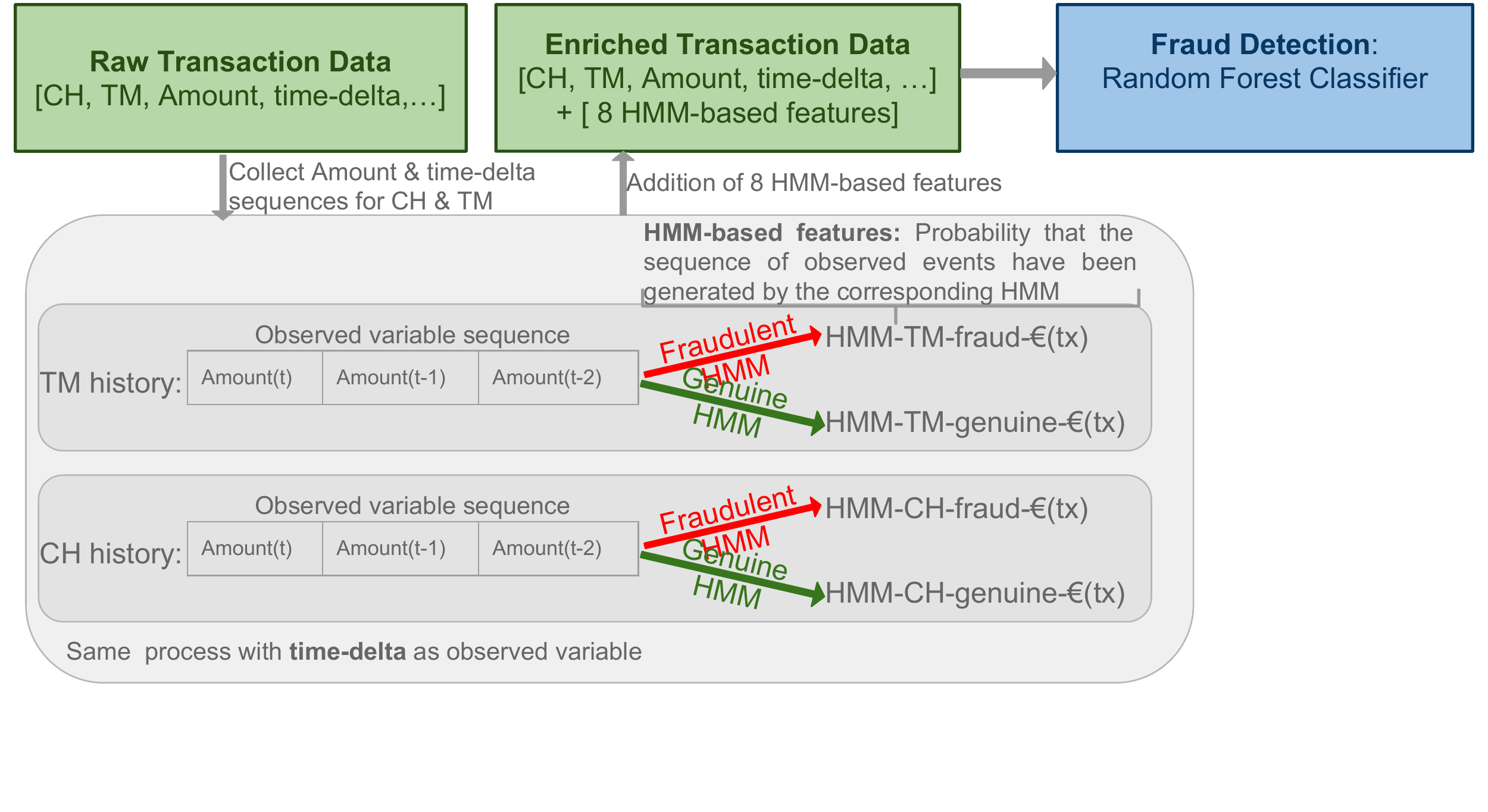}}
\vspace{-1.2 cm}
\caption{Enriching transaction data with HMM-based features calculated from multiple perspectives \textit{(CH$=$Card-holder, TM$=$Terminal)}}
\label{fig_framework}
\end{figure*}

The state of the art feature engineering techniques for credit card fraud detection creates descriptive features using the history of the card-holder (such as: "amount spent by the card-holder in shops from a given country in the last 24h" \cite{whitrow2008} \cite{bahnsen2016}). These descriptive features present several limits we aim to overcome. First they do not take into account the history of the seller even if it is clearly identified in most credit card transactions dataset. Moreover these descriptive features do not consider dependencies between transactions of a same sequence. Therefore we use Hidden Markov Models which are generative probabilistic models and a common choice for sequence modelling \cite{rabiner1991}. %Finally, the choice of the descriptive feature created using the transaction aggregation strategy of Whitrow \& al. \cite{whitrow2008} \cite{bahnsen2016} is guided by expert knowledge. We would like not to be dependent of expert knowledge and we think that doing feature engineering in a supervised way (with the knowledge of the label of transactions in the training set) can be a way to create relevant features without needing expert knowledge about credit card fraud detection.

In addition to the descriptive aggregated features created by Whitrow \& al. \cite{whitrow2008}, we propose to create eight new HMM-based features. They quantify the similarity between the history of a transaction and eight distributions learned previously on set of sequences selected in a supervised way in order to model different perspectives.

In particular, we select three perspectives for modelling a sequence of transactions (see figure \ref{fig_framework}). A sequence (i) can be made only of genuine historical transactions or can include at least one fraudulent transaction in the history, (ii) can come from a fixed card-holder or from a fixed terminal, and (iii) can consist of amount values or of time-delta values (i.e. the difference in time between the current transaction and the previous one). We optimised the parameters of eight HMMs using all eight possible combinations (i-iii). The HMM-based features proposed in this paper are the likelihoods that a sequence is generated by each of these models. 

%\subsection{Multiple perspectives Hidden Markov Models (HMMs)}

In order to make the HMMs model the genuineness and fraudulence of the card holders and the terminals, we create 4 training set containing:
\begin{enumerate}
\item Sequences of transactions from \textbf{genuine credit cards} (without fraudulent transactions in their history).
\item Sequences of transactions from \textbf{genuine terminals} (without fraudulent transactions in their history)
\item Sequences of transactions from \textbf{compromised credit cards} (with at least one fraudulent transaction)
\item Sequences of transactions from \textbf{compromised terminals} (with at least one fraudulent transaction)
\end{enumerate}

We then extract from these sequences of transactions the symbols that will be the observed variable for the HMMs. In our experiments, the observed variable can be either:
\begin{enumerate}
\item The amount of a transaction.
\item The amount of time elapsed between two consecutive transactions of a card-holder (time-delta).
\end{enumerate}

%The transition and emission conditional probability matrices of the Hidden Markov Models are optimized by an iterative Expectation-Maximisation algorithm known as the Baum-Welch algorithm \cite{baum1972} \cite{rabiner1991}. The transition probability matrix is the stochastic matrix that rules the distribution of successive hidden states. In our case, the observed variables are continuous and we make the assumption that they follow a gaussian distribution. The emission probability parameters are the mean and the standard deviation of the gaussian distribution corresponding to each hidden state.

%The choice of Hidden Markov Models over the more recent LSTM and Conditional Random Fields (CBF) was led by the fact that we only want to model the sequence. In our case, the classification is done afterwards by a Random Forest whereas LSTM and CBF are also classifiers. We would like to model the joint probability $P(X,Y)$  with our HMM whereas the classifiers model the conditional probability $P(Y|X=x)$.

At the end, we obtain 8 trained HMMs modeling 4 types of behaviour (genuine terminal behaviour, fraudulent terminal behaviour, genuine card-holder behaviour and fraudulent card-holder behaviour) for both observed variables (amount and time-delta). 

The HMM-based features are the likelihood that the recent sequence of observed events  has been generated by a given HMM.

\section{Experimental Setup}
%\subsection{\textbf{Dataset Description}}
%
%\begin{table}[H]
%\centerline{
%\begin{tabular}{lcc}
%\hline 
% & Fraud & Non-Fraud\\
%\hline
%Face-to-Face & $859$ & $4.1*10^6$\\
%E-commerce & $12493$ & $3.4*10^6$\\
%\hline
%\end{tabular}}
%\caption{Class representation in the test set}
%\end{table} 
%
%The classification task is to predict the class of the transactions (genuine or fraudulent).
%
%Transactions are represented by vectors of continuous, categorical and binary features that characterize the card-holder, the transaction and the terminal. The card-holder is characterized by a unique card-holder ID, its age and its gender. The transaction is characterized by variables like the date-time, the amount and other confidential features. The terminal is characterized by a unique terminal-ID, a merchant category code, a country.
%We also calculate the time-delta feature which is defined as the time spent since the last transaction of the card-holder. %We use this feature for the construction of the proposed HMM-based features

%\subsection{\textbf{Feature Engineering and Dataset Partitioning}}
We use the Python library \textit{hmmlearn}\footnote{\url{https://github.com/hmmlearn/hmmlearn}}. % The number of hidden states was fixed to seven for the results presented in this paper. 

The value of each HMM-based feature is the likelihood that the sequence made of the current transaction and the two previous ones from this terminal/card holder has been generated by the corresponding HMM.  %In our dataset, we could calculate the value of HMM-based features for 80\% of the transactions with this choice of parameter, consequently the remaining 20\% have been deleted.

%We implement six features from the literature on transaction aggregation \cite{whitrow2008} (see table \ref{aggCH}).
%\begin{table}[H]
%\begin{tabular}{ll} 
%Feature & Signification\\
%\hline
%AGGCH1 & $\#$ transactions issued by user in 24h.\\
%AGGCH2 & Amount spent by user in 24h.\\
%AGGCH3 & $\#$ transactions in the country in 24h.\\
%AGGCH4 & Amount spent in the country in  24h.\\
%%AGGCH5 & $\#$ transactions issued by user in the MCC in 24h.\\
%%AGGCH6 & Amount spent by user in the MCC in 24h.\\
%\end{tabular}
%
%\caption{Aggregated features centered on the card-holder(Whitrow \& al. \cite{whitrow2008})}
%\label{aggCH}
%\end{table} 

In order for the HMM-based features and the aggregated features to be comparable, we calculate terminal-centered aggregated features in addition to the card-holder centered aggregated features (see table \ref{aggTM})

\begin{table}[H]
\begin{tabular}{ll}  
Feature & Signification\\
\hline
AGGCH1 & $\#$ transactions issued by user in 24h.\\
AGGCH2 & Amount spent by user in 24h.\\
AGGCH3 & $\#$ transactions in the country in 24h.\\
AGGCH4 & Amount spent in the country in  24h.\\
\hline
AGGTM1 & $\#$ transactions in terminal in 24h.\\
AGGTM2 & Amount spent in terminal in 24h.\\
AGGTM3 & $\#$ transactions with this card type in 24h.\\
AGGTM4 & Amount spent with this card type in 24h.\\
%AGGTM5 & $\#$ transactions in terminal with this card entry mode in 24h.\\
%AGGTM6 & Amount spent in terminal with this card entry mode in 24h.\\
\end{tabular}

\caption{Aggregated features centered on the card holders and the terminal}
\label{aggTM}
\end{table} 	

%In order to quantify the increase in fraud detection when we add HMM-based feature, we compare the Precision-Recall AUC on the test set using Random Forest classifiers trained with different feature sets.

We used a credit card transactions dataset provided by our industrial partner in order to quantify the increase in detection when adding HMM-based features. This dataset contains $4.7*10^7$ anonymized transactions from the belgian credit cards between 01.03.2015 and 31.05.2015. We split temporally the dataset in three different parts: the training set, the validation set and the testing set.% We choose to separate the time period corresponding to the training and validation set and the time period corresponding to the testing set with a gap of 7 days. The reason is that the real world fraud detection systems for human investigators have to verify the alerts generated by the classifiers. Since this process takes time, the ground truth is delayed by about one week. The transactions appearing in this gap of 7 days before the testing set are used in order to calculate the value of the aggregated and HMM-based features in the testing set but not for training the classifiers. Therefore, the training set contains all the transactions between 01.03.2015 to 26.04.2015, the validation set used for the tuning of the random-forests contains all the transactions between 27.04.2015 to 30.04.2015 and the testing set contains all the transactions between 08.05.2015 to 31.05.2015 

We tune the Random Forest hyperparameters through a grid search that optimizes the Precision-Recall Area under the Curve on the validation set. The choice of the Precision-Recall AUC for imbalanced dataset was motivated by the work of Davis \& al. \cite{davis2006}.

\section{Improvement in fraud detection when using HMM-based features}
\label{results}

\begin{figure}[h]
\centerline{
%\vspace{-0.5cm}
\includegraphics[width=0.55\textwidth]{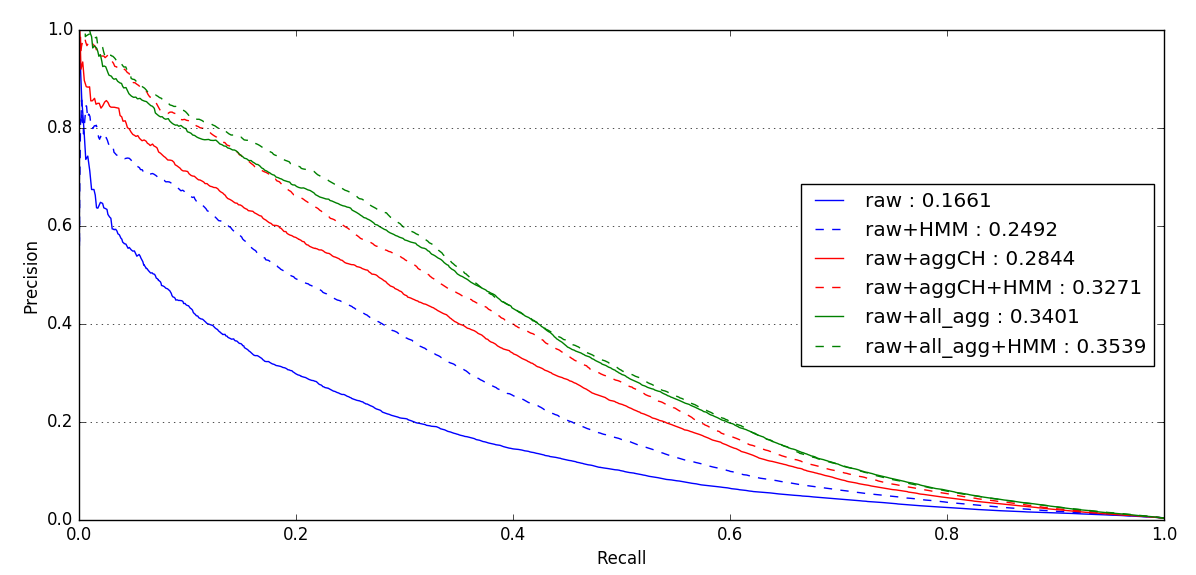}}
\vspace{-0.9 cm}
\caption{Predictions using HMM-based features}%Precision-recall curves for e-commerce transactions.\textit{(Each color corresponds to a specific feature set, the line style corresponds to the presence or absence of HMM-based features. The addition of HMM-based features (dashed lines) to each feature sets - even the most informative ones - allows for an increase in the detection of fraudulent transactions when compared to the same prediction without HMM-based features (solid lines).)}}
\label{ecomresults}
\end{figure}

We train Random Forest Classifiers using different feature sets in order to compare the efficiency of prediction when we add HMM-based features to the classification task.

We tested the addition of our HMM-based features to several feature sets. We refer to the feature set "\textbf{raw+aggCH}" as the state of the art feature engineering strategy since it contains all the raw features with the addition of Whitrow's aggregated features \cite{whitrow2008}. The feature groups we refer to are: the raw features (raw), the features based on the aggregations of card-holders transactions (aggCH), the features based on the aggregation of terminal transactions (aggTM), the proposed HMM-based features (HMM features).

In this section, the HMMs were created with 5 hidden states and the HMM-based features were calculated with a window-size of 3 (actual transaction $+$ 2 past transactions of the card-holder and of the terminal).

By comparing the AUC of the curves raw+aggCH and raw+aggCH+HMM, we observe that adding HMM-based features to the state of the art feature engineering strategy introduced in the work of Whitrow \& al. \cite{whitrow2008} leads to an increase of 15.1\% of the PR-AUC.

The addition of features that describe the sequence of transactions, be it HMM-based features or Whitrow's aggregated features, increases a lot the detection.

\section{Conclusion}

%In this paper, we propose an HMM-based feature engineering strategy that allows us to incorporate sequential knowledge in the dataset in the form of HMM-based features. These HMM-based features enable a non sequential classifier (Random Forest) to use sequential information for the classification. 
The multiple perspective property of our HMM-based feature engineering strategy gives us the possibility to incorporate a broad spectrum of sequential information. In fact, we model the genuine and fraudulent behaviours of the merchants and the card-holders according to two features: the timing and the amount of the transactions. Moreover, the HMM-based features are created in a supervised way and therefore lower the need of expert knowledge for the creation of the fraud detection system.

The results show an increase in the precision-recall AUC of 15.1\% due to the addition of our multi-perspective HMM-based features when compared to the state of the art feature engineering strategies. %We also showed that this increase is robust to the hyperparameters of the HMMs.

HMM-based feature engineering strategy is a powerful tool that is shown to present interesting properties for fraud detection. We can imagine building similar HMM-based features in any supervised task that involve a sequential dataset.

To ensure reproducibility, an optimized code for calculating and evaluating the proposed HMM-based features can be found at \url{https://gitlab.com/Yvan_Lucas/hmm-ccfd} .

\subsection{Acknowledgement:} The work has been funded partially by the Bavarian Ministry of Economic Affairs, Regional Development and Energy in the project “Internetkompetenzzentrum Ostbayern.
\bibliographystyle{splncs}
%\bibliographystyle{apalike}

%\bibliography{sample}

%%----------------------------------------------------------------------------------------

\end{document}